\documentclass[letterpaper, 10 pt, conference]{ieeeconf}

\usepackage{amssymb, amsmath}
\usepackage{graphicx}
\usepackage{subcaption}
\usepackage{epsfig}
\graphicspath{ {./Images/} }

\usepackage{cite}
\usepackage{array}
\usepackage{multirow}
\usepackage[font=small,labelfont=bf]{caption}
\DeclareMathOperator*{\argmin}{argmin} % thin space, limits underneath in displays

\IEEEoverridecommandlockouts                              % This command is only needed if 
\usepackage{color}
\usepackage{algorithm, algorithmic}
\usepackage{tikz}
\tikzstyle{var} = [draw, circle, minimum height=1.2cm,text centered, line width=0.5pt ]
\tikzstyle{block} = [draw,rectangle, rounded corners, minimum width=1cm, minimum height=1cm,text centered, line width=0.5pt ]
\tikzstyle{arrow} = [thick,->,>=stealth,line width=0.5pt]

\title{\LARGE \bf
Online Learning in Planar Pushing with Combined Prediction Model
}

\author{Huidong Gao $^{1*}$, Yi Ouyang $^{2*}$ and Masayoshi Tomizuka$^{1}$% <-this % stops a space
\thanks{$^{1}$Huidong Gao and Masayoshi Tomizuka are with the Department of Mechanical Engineering, University of California, Berkeley, CA 94720 USA
        {\tt\small hgao9@berkeley.edu, tomizuka@berkeley.edu}}%
\thanks{$^{2}$Yi Ouyang is with Preferred Networks, Inc. USA
        {\tt\small ouyangyi@preferred-america.com}}%
\thanks{* indicates equal contributions.}% <-this % stops a space
}

\begin{document}

\maketitle
\thispagestyle{empty}
\pagestyle{empty}

%%%%%%%%%%%%%%%%%%%%%%%%%%%%%%%%%%%%%%%%%%%%%%%%%%%%%%%%%%%%%%%%%%%%%%%%%%%%%%%%
\begin{abstract}
Pushing is a useful robotic capability for positioning and reorienting objects. 
The ability to accurately predict the effect of pushes can enable efficient trajectory planning and complicated object manipulation. 
Physical prediction models for planar pushing have long been established, but their assumptions and requirements usually don't hold in most practical settings.
Data-driven approaches can provide accurate predictions for offline data, but they often have generalizability issues.
In this paper, we propose a combined prediction model and an online learning framework for planar push prediction. 
The combined model consists of a neural network module and analytical components with a low-dimensional parameter. 
We train the neural network offline using pre-collected pushing data. 
In online situations, the low-dimensional analytical parameter is learned directly from online pushes to quickly adapt to the new environments.
We test our combined model and learning framework on real pushing experiments. Our experimental results show that our model is able to quickly adapt to new environments while achieving similar final prediction performance as that of pure neural network models. 

\end{abstract}

%%%%%%%%%%%%%%%%%%%%%%%%%%%%%%%%%%%%%%%%%%%%%%%%%%%%%%%%%%%%%%%%%%%%%%%%%%%%%%%%
\section{Introduction}

Prediction models for physical interactions are important for many robotic tasks such as manipulation, navigation, and motion planning.
Traditionally, models are constructed from analytical methods using physical laws.
However, these analytical models usually fail to provide accurate predictions due to their strong assumptions and requirements for certain physical states and parameters to be known.
Data-driven approaches are getting popular in recent years to build models that predict complex physical dynamics.
They can perform amazingly accurate predictions, but often struggle when transferred to new and unseen situations online. 

In this paper, we investigate the online adaptation abilities of prediction models under new or changing environments. The task we consider is planar pushing.
Pushing is a widely used robotic manipulation action. 
The ability to accurately predict the effect of pushing can enable efficient trajectory planning and complicated object manipulation. 
Although pushing is a simple action, the dynamics are highly non-linear and involve multiple factors such as geometry and mass distribution as well as complex frictional interactions.
Well established analytical models for planar pushing require information about object properties and frictional parameters,\cite{mason1986mechanics, lynch1992manipulation} and their assumptions on frictional forces may not hold. Recent works on using data-driven models for planar pushing improve the push prediction accuracy over analytical models for offline data, but they are not designed for online adaptation \cite{mericcli2015push, kopicki2017learning, zhou2016convex, wang2017focused, bauza2017probabilistic, kloss2017combining,  ajay2018augmenting, byravan2017se3, agrawal2016learning, finn2016unsupervised, xie2019improvisation}.

To achieve quick online adaptation, we propose a combined prediction model and an online learning framework for planar push prediction. 
The combined model consists of a neural network module and several analytical components.
The neural network helps improve the expressiveness of the prediction model, while the
low-dimensional parameter in analytical components allows rapid online adjustment.
We first train the high-dimensional neural network parameter offline using a pre-collected dataset. Then we iteratively learn the low-dimensional analytical parameters online from data collected in the actual push trajectory.
This online learning framework allows us to take advantage of the accuracy of data-driven offline training as well as the rapid online adaptation under new or changing conditions.
We test our combined prediction model and online learning framework in 
two sets of experiments: (1) experiments generated from the MIT Push Dataset\cite{yu2016more}, and (2) real pushing data collected by a modified TurtleBot3 \cite{turtlebot} pushing regular packing boxes. 
The experimental results show that our combined prediction model is able to adapt quickly to unseen situations, and achieves similar final prediction performance compared to offline training loss.

Our main contributions are the following: (1) we provide a novel combined push prediction model consisting of both analytical and data-driven components, (2) we propose an online learning framework where we take advantages of the offline training accuracy while adapting quickly to online situations using the low-dimensional parameter, and (3) we verify the capability of our learning framework with two sets of real robot pushing experiments.

\subsection{Related Work}

Data-driven approaches are popular in pushing prediction, and multiple learning methods have been proposed to train different prediction models.
Gaussian approximation is a basic learning tool which has been used to predict final object poses from initial pushing orientations \cite{mericcli2015push}.
Regular regression and density estimation methods have also been used to train push prediction models \cite{kopicki2017learning} which outperform analytical models.
To further improve predictions on planar pushing, other types of models have been considered in the literature, including physics-based force-motion models \cite{zhou2016convex}, local Markov decision process (MDP) models \cite{wang2017focused}, and heteroscedastic Gaussian process models \cite{bauza2017probabilistic}.

More recently, advances in deep learning have drawn attention to using neural network models for physical interactions.
SE3-Nets \cite{byravan2017se3} use deep neural networks to
directly predict rigid body motions from point cloud data.
Image prediction based deep learning models  \cite{agrawal2016learning, finn2016unsupervised, xie2019improvisation} also show success in several manipulation tasks including planar pushing.
These deep learning models generally improve the prediction accuracy on specific distributions of the given dataset, but they often have generalizability issues for unseen data distributions.

One approach to improve the transferability of neural networks is to use hybrid architectures that combine analytical and data-driven models \cite{kloss2017combining, ajay2018augmenting}. This approach benefits from the expressiveness of data-driven models and the generalizability of analytical methods.
Our combined push prediction model is inspired by the generalization abilities of these hybrid models. We further push beyond generalization to online adaptation and design models that take advantages of both offline and online training.

There are other deep learning based online adaptation methods using either recurrent neural networks \cite{li2018push} or meta learning \cite{nagabandi2018deep}. Although these methods seem appealing, they require complex training procedures which often can only be done in simulators.
In comparison, our combined model and online learning method provide a simple though effective online adaptation scheme.

% With control: 
% \begin{itemize}
%     \item Example: Data-efficient control paper \cite{bauza2018data}: shows can find effective control policies with small data to achieve accurate closed loop performance. Only compares analytical and pure data driven controller within a MPC structure, no hybrid. 
%     \item Example: Combining Analytical and Learned Models for Model Predictive Control Paper \cite{baumeister2018combining}: uses hybrid model within MPC, learn h with FCC layer. Did not consider center of mass(COM), only uses MPC for online control but not online model adjustment, also it doesn't test on real experiment data.
% \end{itemize}

%%%%%%%%%%%%%%%%%%%%%%%%%%%%%%%%%%%%%%%%%%%%%%%%%%%%%%%%%%%%%%%%%%%%%%%%%%%%%%%%

\section{Preliminaries}

\subsection{Planar Push Prediction}

We consider the prediction problem of the pushing behavior between a robot and an object on a surface.
At each time $t$, we observe the (center) position $\mathbf p_o(t) = (\mathbf p_{o, x}(t), \mathbf p_{o, y}(t)) \in \mathbb R^2$ and orientation $\omega_o(t) \in [0, 2\pi)$ of the object,
the position $\mathbf p_r(t) = (\mathbf p_{r, x}(t), \mathbf p_{r, y}(t)) \in \mathbb{R}^2$ of the robot, and the robot's motion command $\mathbf u_r(t) = (\mathbf u_{r,x}(t), \mathbf u_{r,y}(t)) \in \mathbb R^2$.
The robot will then move according to $\mathbf u(t)$ to $\mathbf p_t(t+1) = \mathbf p_r(t) + \mathbf u_r(t)$ and push the object along its way.
Given the observation, our goal is to predict the object's next position $\mathbf p_o(t+1)$ and orientation $\omega_o(t+1)$ at time $t+1$ after pushed by the robot.

\begin{figure}[h]
\begin{center}
\begin{tikzpicture}[scale=0.8, every node/.style={scale=0.8}]
\def\pox{0}
\def\poy{0}
\def\width{3.6}
\def\height{2}

\def\prx{1}
\def\pry{-2}
\pgfmathsetmacro\radius{-\pry-\height/2}

\def\ux{1}
\def\uy{1}

\def\vx{0.7}
\def\vy{1}

\def\dpx{0.5}
\def\dpy{0.8}

\node[draw, rectangle,minimum width=\width cm, minimum height=\height cm, line width=0.5pt ] 
        at (\pox, \poy) (box) {};
\node[circle, fill, scale=.5, label=left:{$\mathbf p_o$}] at (\pox, \poy) (po) {};

\node[draw, circle, minimum height=2*\radius cm, line width=0.5pt] 
        at (\prx,\pry) (robot) {}; 
\node[circle, fill, scale=.5] at (\prx, \pry) (pr) {};

\draw[arrow] (po) --node[left, pos=0.85] {$\mathbf p_r^o$} (pr);

\node[circle, fill, scale=.5] at (\prx, \pry + \radius) (c) {};

\draw[arrow] (po) --node[right, pos=0.75] {$\mathbf c$} (c);

\node at (\prx + \ux, \pry + \uy) (u) {};
\draw[arrow] (pr) --node[below]{$\mathbf u_r^o$} (u);

\node at (\prx + \vx, \pry + \radius + \vy) (v) {};
\draw[arrow] (c) --node[right]{$\mathbf u_c$} (v);

\node at (\pox + \dpx, \poy + \dpy) (dp) {};
\draw[arrow, dashed] (po) --node[left]{$\Delta \mathbf p_o$} (dp);

\node at (\pox+\width/2, \poy+\height/2) (w) {};
\draw[arrow, dashed] (w.120) arc (0:50:1cm) node[left, midway] {$\Delta \omega_o$};

\node[draw, rectangle, minimum width=\width cm, minimum height=\height cm, line width=0.5pt,
        dashed, rotate around={15:(dp)}] 
        at (dp) (boxrotated) {};

\end{tikzpicture}
\caption{Planar Pushing}
\label{fig:pushing}
\end{center}
\end{figure}
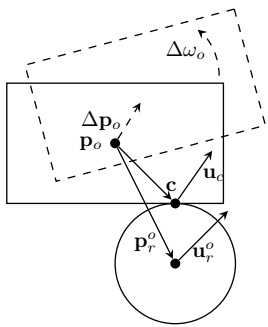

Since the goal is to predict the object's movement, we transform the observations and predictions into the object's current frame. Specifically, let $\mathbf p^o_r(t)$ and $\mathbf u_r^o(t)$ denote the relative position and motion of the robot, and $\Delta \mathbf p_o(t)$ and $\Delta \omega_o(t)$ be the relative change in position and orientation of the object. They satisfy the following equations:
\begin{align}
    & \mathbf p^o_r(t) = \mathbf{R}_{(- \omega_o(t))}  (\mathbf p_r(t) - \mathbf p_o(t)) 
    \\
    & \mathbf u_r^o(t) = \mathbf{R}_{(- \omega_o(t))} \mathbf u_r(t)
    \\
    & \Delta p_o(t) = \mathbf{R}_{(- \omega_o(t))}  (\mathbf p_o(t + 1) - \mathbf p_o(t)) 
    \\
    & \Delta \omega_o(t) = \omega_o(t+1) - \omega_o(t)
\end{align}
where $\mathbf{R}_{\omega}$ denotes the rotation matrix by the angle $\omega$.

Figure \ref{fig:pushing} illustrates the planar pushing interaction between the robot and the object. For notation simplicity, we will drop the time index $t$ if it's clear in the context in the rest of the paper.
Then the goal of push prediction is to find a prediction function $f_\theta$, parameterized by $\theta$, that predicts
$f_\theta ( \mathbf p^o_r , \mathbf u_r^o) 
    = ( \hat{\Delta} \mathbf p_o, \hat{\Delta} \omega_o)$.

\subsection{Performance Metric}

To evaluate the prediction performance, we use the standard metric of normalized mean square error (NMSE) for both positional and rotational losses. The same metric was also used in data-driven pushing models \cite{bauza2017probabilistic}.

To compute NMSE, we define positional loss functions $\ell_{\text{pos},x}$, $\ell_{\text{pos},y}$, and a rotational loss function $\ell_{\text{rot}}$ by
\begin{align}
    & \ell_{\text{pos}, i} (\hat{\Delta} \mathbf p_o, \Delta {\mathbf p}_o )  =
     \frac{1}{\sigma_{\Delta \mathbf p_o}^2} (\hat{\Delta} \mathbf p_{o,i} - \Delta \mathbf p_{o,i})^2, \, i = x, y
    % \\
    % & \ell_{\text{pos}, y} (\hat{\Delta} \mathbf p_o, \Delta {\mathbf p}_o )  =
    % \frac{1}{\sigma_{\Delta \mathbf p_o}^2} (\hat{\Delta} \mathbf p_{o,y} - \Delta \mathbf p_{o,y})^2 
    \\
    & \ell_{\text{rot}}  ( \hat{\Delta} \omega_o ,\Delta \omega_o) 
 = \frac{1}{\sigma_{\Delta \omega_o}^2}(\hat{\Delta} \omega_o - \Delta \omega_o)^2
\end{align}
where $\sigma_{\Delta \mathbf p_o}$ and $\sigma_{\Delta \omega_o}$ are the standard deviations for $\Delta \mathbf p_o$ and $\Delta \omega_o$.
Positional and rotational NMSEs are given by
\begin{align}
    &\text{NMSE}_{\text{pos}} \! =\! \frac{1}{2}\mathbb E \big[
    \ell_{\text{pos}, x} (\hat{\Delta} \mathbf p_o, \Delta {\mathbf p}_o )
    \! + \! \ell_{\text{pos}, y} (\hat{\Delta} \mathbf p_o, \Delta {\mathbf p}_o ) 
    \big]
    \\
    & \text{NMSE}_{\text{rot}} = \mathbb E \big[
    \ell_{\text{rot}}  ( \hat{\Delta} \omega_o ,\Delta \omega_o) \big]
\end{align}
We also define the overall loss function $\ell$ as
\begin{align}
    &\ell((\hat{\Delta} \mathbf p_o, \hat{\Delta} \omega_o),\, (\Delta {\mathbf p}_o, \Delta \omega_o) ) 
    = \ell_{\text{pos}, x} (\hat{\Delta} \mathbf p_o, \Delta {\mathbf p}_o )
    \notag\\
    & \qquad + \ell_{\text{pos}, y} (\hat{\Delta} \mathbf p_o, \Delta {\mathbf p}_o )
    + \ell_{\text{rot}}  ( \hat{\Delta} \omega_o ,\Delta \omega_o) 
\end{align}
% In order to make the prediction invariant to object orientation, we describe $\mathbf{R}_{pos}^t$ in object frame at time $t$ according to equation (1). We also change the output at time $t+1$ to be the relative object positional and rotational movements to object frame at time $t$.

% \begin{equation}
% \mathbf{R}_{p_{rel}}^t = \mathbf{T}_{- \mathbf{O}_{rot}^t}  \cdot (\mathbf{R}_{pos}^t - \mathbf{O}_{pos}^t) 
% \end{equation}

% \begin{equation}
% \mathbf{O}_{p_{rel}}^{t+1} = \mathbf{T}_{- \mathbf{O}_{rot}^{t}} \cdot (\mathbf{O}_{pos}^{t+1} - \mathbf{O}_{pos}^{t}) 
% \end{equation}

% \begin{equation}
% \mathbf{O}_{r_{rel}}^{t+1} = (\mathbf{O}_{rot}^{t+1} - \mathbf{O}_{rot}^{t})
% \end{equation}

% where $\mathbf{R}_{p_{rel}}^t$ denotes relative position of robot at time $t$, $\mathbf{T}_{- \mathbf{O}_{rot}^t}$ denotes rotation matrix that rotates given vector by an angle of $-\mathbf{O}_{rot}^{t}$. $\mathbf{O}_{p_{rel}}^{t+1}$ and $\mathbf{O}_{r_{rel}}^{t+1}$ denotes the relative object positional and rotational movements from $t$ to $t+1$, respectively.

\subsection{Physical Model}

Suppose the center of mass (COM) of the object is at the object center such that $\mathbf p_o = \text{COM} = (\text{COM}_x, \text{COM}_y)$. Let $\mathbf c = (\mathbf c_x, \mathbf c_y)$ be the pushing contact point in the object's frame, and 
$\mathbf u_c = (\mathbf u_{c,x}, \mathbf u_{c,y})$ is the motion (in the object's frame) of the contact point being pushed by the robot.
When $\mathbf c$, $\mathbf u_c$ and a friction-related parameter $h$ are available, the physical model of pushing dynamics \cite{lynch1992manipulation} gives
\begin{align}
    & \Delta\text{COM}_x = 
    \frac{ (h^2 + \mathbf c_x^2) \mathbf u_{c,x} + \mathbf c^o_x \mathbf c^o_y \mathbf u_{c,y}}
    {h^2 + \mathbf c_x^2 + \mathbf c_y^2 }
    \label{eq:push_dynamics_x}
    \\
    & \Delta\text{COM}_y = 
    \frac{ (h^2 + \mathbf c_y^2) \mathbf {c,y} + \mathbf c_x \mathbf c_y \mathbf {c,x} }
    {h^2 + \mathbf c_x^2 + \mathbf c_y^2 }
    \label{eq:push_dynamics_y}
    \\
    & \Delta\omega_o= \frac{ \mathbf c_x \Delta\text{COM}_y - \mathbf c_y \Delta\text{COM}_x}{h^2}
    \label{eq:push_dynamics_omega}
\end{align}

We use $F_{\text{physical}}$ to denote this physical model \eqref{eq:push_dynamics_x}-\eqref{eq:push_dynamics_omega} as
$F_{\text{physical}}(\mathbf c, \mathbf u_c) = (\Delta\text{COM}, \Delta\omega_o)$.
One may attempt to directly use this model as a prediction function, but the physical model is subjected to several limitations:
(1) COM of the object may not be at its center $\mathbf p_o$. For example, when the object is a box containing items with different weights, its COM is usually different from its geometric center.
(2) It's difficult to determine the exact contact point $\mathbf c$ between the robot and the object from their positions. Moreover, a push can be either sticking or slipping. For a sticking push we can simply get $\mathbf u_c = \mathbf u_r$. However, in the slipping case, $\mathbf u_c$ will depend on the complex friction interaction between the robot and the object. 
(3) The physical model requires the knowledge of a friction-related parameter $h$, but this parameter varies for different contact surfaces and is usually unknown for unseen objects.

% \begin{enumerate}
%     \item COM of the object may not be at its center $\mathbf p_o$. For example, when the object is a box with multiple items of different weights inside, its COM is usually different from its geometric center.
%     \item It's difficult to determine the exact contact point $\mathbf c$ between the robot and the object from their positions. Moreover, a push can be either sticking or slipping. For a sticking push we can simply get $\mathbf u_c = \mathbf u_r$. However, in the slipping case, $\mathbf u_c$ will depend on the complex friction interaction between the robot and the object. 
    
%     \item The physical model requires the knowledge of a friction-related parameter $h$, but this parameter varies for different contact surfaces and is usually unknown for an unseen object.
% \end{enumerate}
Therefore, directly applying the physical model may result in inaccurate predictions due to above issues.

%%%%%%%%%%%%%%%%%%%%%%%%%%%%%%%%%%%%%%%%%%%%%%%%%%%%%%%%%%%%%%%%%%%%%%%%%%%%%%%%

\section{Learning Method and Prediction Model}

\subsection{Online Learning for Planar Push Prediction}

When a pushing dataset is available, data-driven approaches can learn prediction functions offline.
Offline trained functions may perform well for situations similar to the collected dataset, but they usually have generalizability issues with unseen cases.

In planar pushing, many important factors can vary in different scenarios. For example, different objects have different weights, friction coefficients, and different COM positions.
Furthermore, these pushing-related factors are often not available to the robot before it actually pushes the object.
In most cases, the only way to infer these properties is to observe the online pushing results. Therefore, online learning is essential to perform accurate push predictions.

Taking advantages of both online and offline training, we consider an online learning setting with offline pre-training.
In particular, we split the prediction function parameter into $\theta = (\theta_{\text{offline}}, \theta_{\text{online}})$.
The offline component $\theta_{\text{offline}}$ is trained offline with a pre-collected dataset while
the online component $\theta_{\text{online}}$ will be learned online to adapt to new pushing trajectories.
The idea is that the offline component $\theta_{\text{offline}}$ can be high-dimensional to improve the expressiveness of the prediction model. On the other hand, the online component $\theta_{\text{online}}$ can be designed to be low-dimensional for fast online adaptation.

Suppose we have a dataset $D$ consisting of data points of the form $(x,y)$ where $x = (\mathbf p^o_r , \mathbf u^o)$ and $y = ( \Delta {\mathbf p}_o, \Delta \omega_o)$.
Then the goal of the offline pre-training phase is to find optimal offline parameter
\begin{align}
    \theta_{\text{offline}}^*  = \argmin_{\theta_{\text{offline}}} 
    \frac{1}{|D|} \sum_{(x,y) \in D} \ell ( f_{(\theta_{\text{offline}}, \theta_{\text{online}}(0))}(x), y)
    \label{eq:offline_training}
\end{align}
Note that we fix the online parameter to an initial value $\theta_{\text{online}}(0)$ in offline training.
An optional scheme is to also train the online parameter $\theta_{\text{online}}$ offline, but this may require additional hyperparameter tuning.

When it comes to the online situation, we have a pre-trained prediction function
$f_{(\theta_{\text{offline}}^*, \theta_{\text{online}}(0))}$ at time $0$. 
The main idea of online learning is to adjust the online parameter $\theta_{\text{online}}(t)$ adapting to the pushing outcomes at each time $t$.

Let $x(t) = (\mathbf p^o_r(t) , \mathbf u^o(t))$ and $y(t) = ( \Delta {\mathbf p}_o(t), \Delta \omega_o(t))$ be the pushing outcome to be predicted.
Then the push prediction at this time is $f_{(\theta_{\text{offline}}^*, \theta_{\text{online}})} ( x(t) )$ with a prediction loss
$\ell ( f_{(\theta_{\text{offline}}^*, \theta_{\text{online}}(t))} ( x(t) ), y(t) )$.

After the pushing outcome $y(t)$ is observed at time $t+1$, we can update the online parameter by
\begin{align}
    \theta_{\text{online}} (t+1)  = \argmin_{\theta_{\text{online}}} 
    \ell ( f_{(\theta_{\text{offline}}^*, \theta_{\text{online}})}(x(t)), y(t))
    \label{eq:online_training}
\end{align}

The overall online learning algorithm is described below.
\begin{algorithm}[H]
\caption{Online Learning for Push Prediction}
\label{alg:overall}
\begin{algorithmic}
\STATE \textbf{Inputs}: 
\STATE \hspace{\algorithmicindent} Dataset $D$ of pushing trajectories
\STATE \hspace{\algorithmicindent} Prediction model $f_{(\theta_{\text{offline}}, \theta_{\text{online}})}$
\STATE \hspace{\algorithmicindent} Initial online parameter $\theta_{\text{online}}(0)$
\STATE \hspace{\algorithmicindent} Loss function $\ell$

\STATE Train offline parameter $\theta_{\text{offline}}^*$ by \eqref{eq:offline_training}

\FOR{$t = 0, 1, 2, \dots$}
    \STATE Observe $x(t)$ and $y(t)$ online
    \STATE Update online parameter $\theta_{\text{online}}(t+1)$ by \eqref{eq:online_training}
    
\ENDFOR{}
\end{algorithmic}
\end{algorithm}

To apply this online learning framework, we need to have a prediction model $f_{(\theta_{\text{offline}}, \theta_{\text{online}})}$ with a high-dimensional offline parameter $\theta_{\text{offline}}$ and a low-dimensional online parameter $\theta_{\text{online}}$.
To do this, we will make use of analytical models as low-dimensional online components, and a data-driven model for the high-dimensional offline component.
These components will be constructed in the next subsections.

%%%%%%%%%%%%%%%%%%%%%%%%%%%%%%%%%%%%%%%%%%%%%%%%%%%%%%%%%%%%%%%%%%%%%%%%%%%%%%%%

\subsection{Center of Mass Corrections}
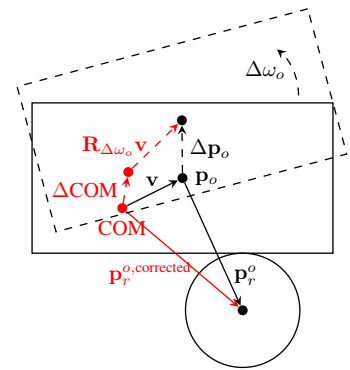
\begin{figure}[h]
\begin{center}
\begin{tikzpicture}[scale=0.8, every node/.style={scale=0.8}]
\def\pox{0}
\def\poy{0}
\def\width{5}
\def\height{2.5}

\pgfmathsetmacro\comx{\pox-1.0}
\pgfmathsetmacro\comy{\poy-0.5}

\def\prx{1}
\def\pry{-2.2}
\pgfmathsetmacro\radius{-\pry-\height/2}

\node[draw, rectangle,minimum width=\width cm, minimum height=\height cm, line width=0.5pt ] 
        at (\pox, \poy) (box) {};
\node[circle, fill, scale=.5, label=right:{$\mathbf p_o$}] at (\pox, \poy) (po) {};

\node[circle, fill, scale=.5, label={[red]270:{COM}}, red] at (\comx, \comy) (com) {};

\draw[arrow] (com) --node[above] {$\mathbf v$} (po);

\node[draw, circle, minimum height=2*\radius cm, line width=0.5pt] 
        at (\prx,\pry) (robot) {}; 
\node[circle, fill, scale=.5] at (\prx, \pry) (pr) {};

\draw[arrow] (po) --node[right, pos=0.75] {$\mathbf p_r^o$} (pr);

\draw[arrow, red] (com) --node[left, pos=0.65] {$\mathbf p_r^{o, \text{corrected}}$} (pr);

\def\dpx{0.1}
\def\dpy{0.6}

\node[circle, fill, scale=.5, red] at (\comx + \dpx, \comy + \dpy) (dcom) {};
\draw[arrow, dashed, red] (com) --node[left]{$\Delta$COM } (dcom);

\node[circle, fill, scale=.5] at (\comx + \dpx + 0.89, \comy + \dpy + 0.86) (dp) {};

\draw[arrow, dashed, red] (dcom) --node[left]{$\mathbf R_{\Delta \omega_o} \mathbf v$} (dp);

\draw[arrow, dashed] (po) --node[right]{$\Delta \mathbf p_o$} (dp);

\node[draw, rectangle, minimum width=\width cm, minimum height=\height cm, line width=0.5pt,
        dashed, rotate around={15:(dp)}] 
        at (dp) (boxrotated) {};
\node at (\pox+\width/2-0.5, \poy+\height/2) (w) {};
\draw[arrow, dashed] (w.120) arc (0:50:1cm) node[left, midway] {$\Delta \omega_o$};
\end{tikzpicture}
\caption{Center of Mass Corrections}
\label{fig:com_correction}
\end{center}
\end{figure}

One challenge in push prediction is to know the center of mass (COM) of the object. The object center $\mathbf p_o$ can be an approximation of COM, but it usually has an offset from COM in many situations.
Therefore, we propose an analytical procedure to correct COM as shown in Figure \ref{fig:com_correction}.

Consider $\omega_o = 0$ without loss of generality. Let $\mathbf v = \mathbf p_o - \text{COM} \in \mathbb{R}^2$ be the offset vector from the true COM to the object center in the object's frame. Then the robot's relative position to the object should be corrected to its relative position to the COM by
\begin{align}
    \mathbf p_r^{o, \text{corrected}} = 
    \mathbf p_r - \text{COM} = \mathbf p_r^o + \mathbf v
    \label{eq:position_correction}
\end{align}
For the push prediction, note that the offset between the object center and COM after pushing is the rotated vector $\mathbf R_{\Delta \omega_o} \mathbf v$.
Suppose the COM motion is $\Delta$COM after being pushed, then the motion of the object center is given by
\begin{align}
    \Delta \mathbf p_o = & \mathbf p_o(t+1) - \mathbf p_o(t)
                           \notag \\
                          = & \text{COM}(t+1) + R_{\Delta \omega_o} \mathbf v - \mathbf p_o(t)
                          \notag \\
                          = & \Delta\text{COM} + (R_{\Delta \omega_o} - \mathbf I) \mathbf v
                          \label{eq:motion_correction}
\end{align}

\subsection{Contact Point Prediction for Physical Model}
Since we want to apply the physical model $F_{\text{physical}}$, we need to predict the contact point $\mathbf c$ and the contact motion $\mathbf u_c$.
As discussed earlier, predicting the contact point and motion is one of the key challenges in analyzing pushing behaviors. Therefore, we take advantage of data-driven ideas to train a model for the complex contact interactions.
We achieve this by using a feedforward neural network $\phi_{\theta_{\text{offline}}}$ parameterized by an offline parameter $\theta_{\text{offline}}$.
In particular, this neural network will take the corrected relative robot position and motion as input, and output the predicted contact point and motion
$\phi_{\theta_{\text{offline}}}(\mathbf p_r^{o, \text{corrected}}, \mathbf u_r^o) = (\mathbf c, \mathbf u_c)$.

\subsection{Combined Push Prediction Model}

As shown in Figure \ref{fig:com_correction}, putting together the neural network $\phi_{\theta_{\text{offline}}}$, physical model $F_{\text{physical}}$, and COM corrections \eqref{eq:position_correction}-\eqref{eq:motion_correction} we get the combined push prediction model:
\begin{align}
    f_{(\theta_{\text{offline}}, \theta_{\text{online}})}(\mathbf p^o_r, \mathbf u^o_r) = (\Delta \mathbf p_o, \Delta \omega_o)
\end{align}

In the prediction model, we have a high-dimensional offline parameter $\theta_{\text{offline}}$ from the neural network which can improve the expressive power of the model. Form the analytical components $F_{\text{physical}}$ and COM corrections \eqref{eq:position_correction}-\eqref{eq:motion_correction}, we have a low-dimensional online parameter $\theta_{\text{online}} = (\mathbf v, h) \in \mathbb{R}^3$.
Since the online parameter $\theta_{\text{online}}$ has a low dimension, it can be quickly trained to adapt to online data.

One important features of the model is that all the components are differentiable. This allows us to perform both offline and online training in an end-to-end manner.
Note that it's commonly observed in deep models that end-to-end training can further exploit the expressive power by letting the network to determine its own state representation.
This also implies that each intermediate variable may be trained to behave differently than its analytical role.

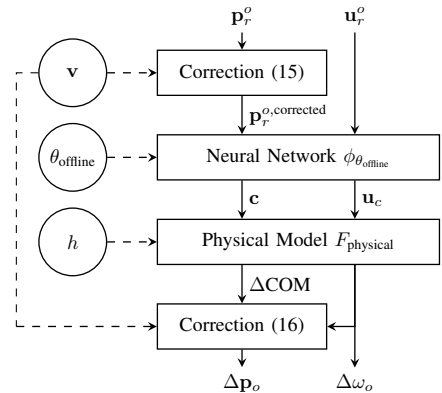
\begin{figure}
\begin{center}
\begin{tikzpicture}[x=1cm,y=1cm, scale=0.75, every node/.style={scale=0.75}]
\node at (0, 0) (pr) {$\mathbf p^o_r$};
\node at (2, 0) (ur) {$\mathbf u^o_r$};

\node[draw, rectangle, minimum width=3 cm, minimum height=0.8 cm, line width=0.5pt ] 
        at (0, - 1) (cominput) {Correction \eqref{eq:position_correction}};

\node[draw, rectangle, minimum width=5 cm, minimum height=0.8 cm, line width=0.5pt ] 
        at (1 , -2.5) (NN) {Neural Network $\phi_{\theta_{\text{offline}}}$};

\node[draw, rectangle, minimum width=5 cm, minimum height=0.8 cm, line width=0.5pt ] 
        at (1, -4) (physical) {Physical Model $F_{\text{physical}}$};

\draw[arrow] (pr) -- (cominput);
\draw[arrow] (cominput.south) --node[right]{$\mathbf p_r^{o, \text{corrected}}$} (pr.south|-NN.north);
\draw[arrow] (ur) -- (ur.south|-NN.north);

\draw[arrow] (pr.south|-NN.south) --node[right]{$\mathbf c$} (pr.south|-physical.north);
\draw[arrow] (ur.south|-NN.south) --node[right]{$\mathbf u_c$} (ur.south|-physical.north);

\node at (0, -6.5) (dpo) {$\Delta \mathbf p_o$};
\node at (2, -6.5) (domega) {$\Delta \omega_o$};

\node[draw, rectangle, minimum width=3 cm, minimum height=0.8 cm, line width=0.5pt ] 
        at (0, - 5.5) (comoutput) {Correction \eqref{eq:motion_correction}};

\draw[arrow] (comoutput) -- (dpo);
\draw[arrow] (physical.south-|comoutput.north) --node[right]{$\Delta$COM} (comoutput.north);
\draw[arrow] (domega.north|-physical.south) -- (domega.north);

\draw[arrow] (domega.north|-physical.south) |- (comoutput.east);

\node[draw, circle, minimum height=1.2 cm, line width=0.5pt ] 
        at (-3, - 2.5) (thetaoff) {$\theta_{\text{offline}}$};

\node[draw, circle, minimum height=1.2 cm, line width=0.5pt ] 
        at (-3, - 4) (h) {$h$};

\node[draw, circle, minimum height=1.2 cm, line width=0.5pt ] 
        at (-3, - 1) (v) {$\mathbf v$};

\draw[arrow, dashed] (thetaoff) -- (NN);
\draw[arrow, dashed] (h) -- (physical);
\draw[arrow, dashed] (v) -- (cominput);
\draw[dashed] (v) -| (-4, - 5.5);
\draw[arrow, dashed] (-4, - 5.5) -- (comoutput);

\end{tikzpicture}
\caption{Combined Push Prediction Model $f_{(\theta_{\text{offline}}, \theta_{\text{online}})}$}
\label{fig:combined_model}
\end{center}
\end{figure}

%%%%%%%%%%%%%%%%%%%%%%%%%%%%%%%%%%%%%%%%%%%%%%%%%%%%%%%%%%%%%%%%%%%%%%%%%%%%%%%%

\section{Experiments}

To test our online learning algorithm and the combined push prediction model, we consider two different sets of experiments. 
The first set of experiments are based on data from the MIT push dataset \cite{yu2016more}. The MIT dataset is collected by a robot arm with a stiff cylindrical steel pusher ($9.5$ mm diameter) pushing a variety of small objects (about $100$ mm wide) on four different surfaces.
To simulate the online situation, we create a simulator that outputs one data point at a time from a complete trajectory. The model is then updated online with the simulated pushing trajectory.

In the second set of experiments, we collected real data using a TurtleBot3 \cite{turtlebot}. Since our focus is on predicting positional and rotational changes by planar point pushing, we modify the shape of the robot to a disc by covering it with a round basin (radius $350$ mm).  The experiments include pushing boxes of two different sizes, different COM positions, and with different contact surfaces with the floor.

The two sets of experiments vary in data collection frequency, pusher size and object type, in order to show our method works under a variety of settings.
Since our goal is to demonstrate the online learning ability of the combined push prediction model, in all the experiments we choose a simple neural network architecture consisting of three fully connected layers, with each layer having $16$ units followed by the ReLU activation.
The entire combined model, including the analytical components and the neural network, is implemented in the deep learning framework Chainer \cite{chainer_learningsys2015}. 
Before model training, each of the input and output variable to the prediction model is normalized to zero mean and unit variance.
In the offline training phase, we use the Adam optimizer \cite{kingma2014adam} with learning rate $0.005$ and batch size $32$.
In online learning, at every time step the online optimization problem \eqref{eq:online_training} is solved by gradient decent with $5$ gradient steps with learning rate $0.005$. 

%% explain captions here
For all the experiment figures, the caption: setting1/setting2, setting3/setting4, setting5/setting6 (if there is a third row) means that the first row's offline setting is setting1, online setting is setting2, and similarly for the second and the third rows.
The x-axis represents the number of time steps, and y-axis represents online losses at each step. 
Subfigures in the left column show the total loss with mean and shaded one standard deviation area. Subfigures in the right column show the mean values of positional (pos) and rotational (rot) losses.
The blue and orange curves represent results of the fixed (offline trained) model and the model with online learning. The black lines indicate the average offline training losses.
We show the curves with a moving average of 10 time steps for better visualization.
The offline training losses and average online losses for fixed and online learning models are shown in Tables \ref{table:MIT} and \ref{table:turtlebot} where we also add the prediction losses of a pure neural network model trained offline as a baseline.
Note that the offline NMSE of our model is of similar magnitude as that of existing data-driven models \cite{bauza2017probabilistic} for the MIT dataset.

\subsection{MIT Push Dataset}
    The experiments are conducted as following. For a certain object, material, COM and pushing side setting, a total of around 20 straight push trajectories are used to test online learning. The 20 trajectories consist of different pushing points and pushing angles. For offline training, each setting has around 50 straight push trajectories. Each offline and online trajectory contains around 500 data points, collected at 250 Hz.The cases we consider include: Three objects: rect1, rect2, and rect3. Four materials: abs, delrin, plywood, pu. Two COM positions: center, UR $ = (0.01m, 0.01m)$. Pushing sides: front, left. See \cite{yu2016more} for more details about the dataset.
    % \begin{itemize}
    %     \item Three objects: rect1, rect2, and rect3. 
    %     \item Four materials: abs, delrin, plywood, pu. 
    %     \item Two COM positions: center, $(0.01, 0.01)$ m
    %     \item Pushing sides: front, left 
    % \end{itemize}
    \hspace{1mm}
    \subsubsection{Exp M1: different materials }
    We consider rec2 object and do offline and online training with different materials. The goal is to verify if our model is able to adjust for different friction friction coefficients online. 
    Note that in the top case the online setting also appears in offline data. However, online learned model still achieve much better performance over the fixed model.
    \begin{figure}
    \centering
    \includegraphics[width=0.48\textwidth]{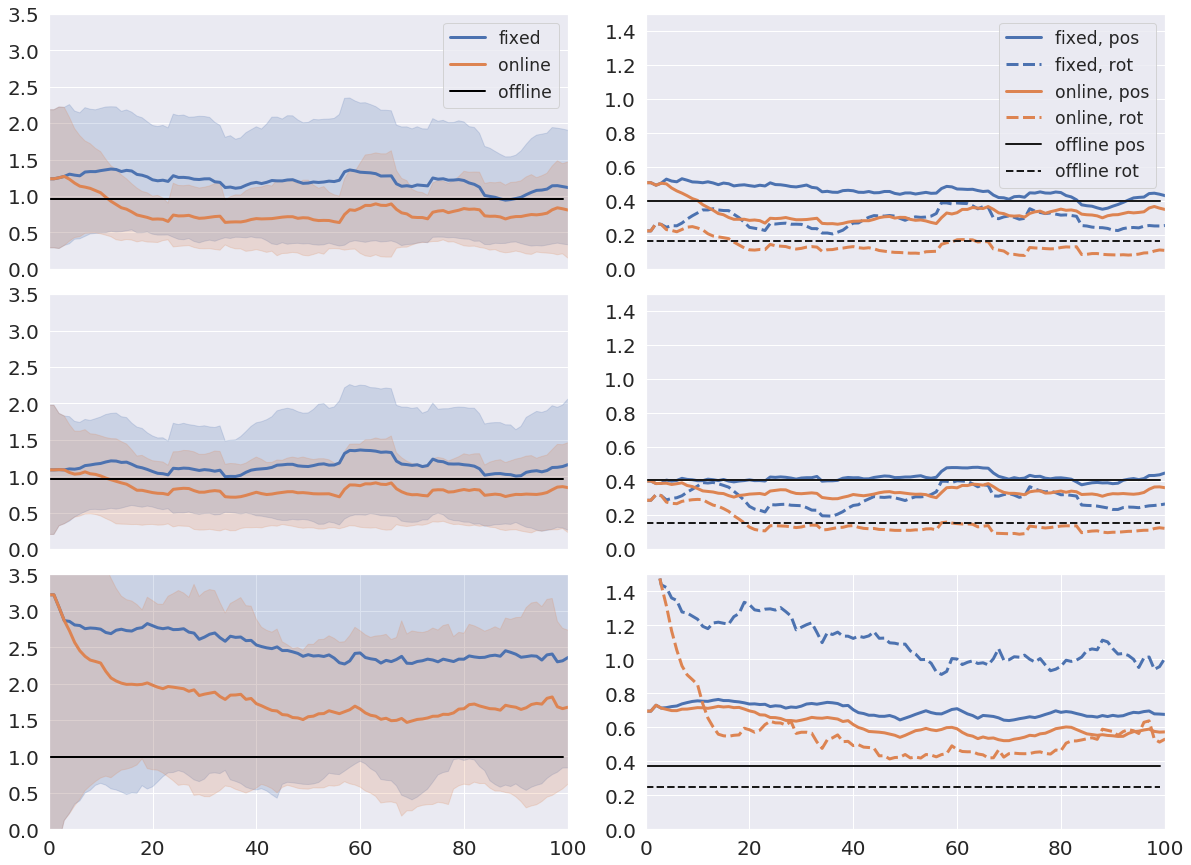}
    \caption{
    Exp M1. (delrin+abs)/abs, delrin/abs, plywood/pu
    }
    \end{figure}
    
    \subsubsection{Exp M2: manual online COM offsets }
    We consider the rec2 object and add a manual COM offset of $(0.01m, 0.01m)$ to the object position online.
    
    \begin{figure}
    \centering
    \includegraphics[width=0.48\textwidth]{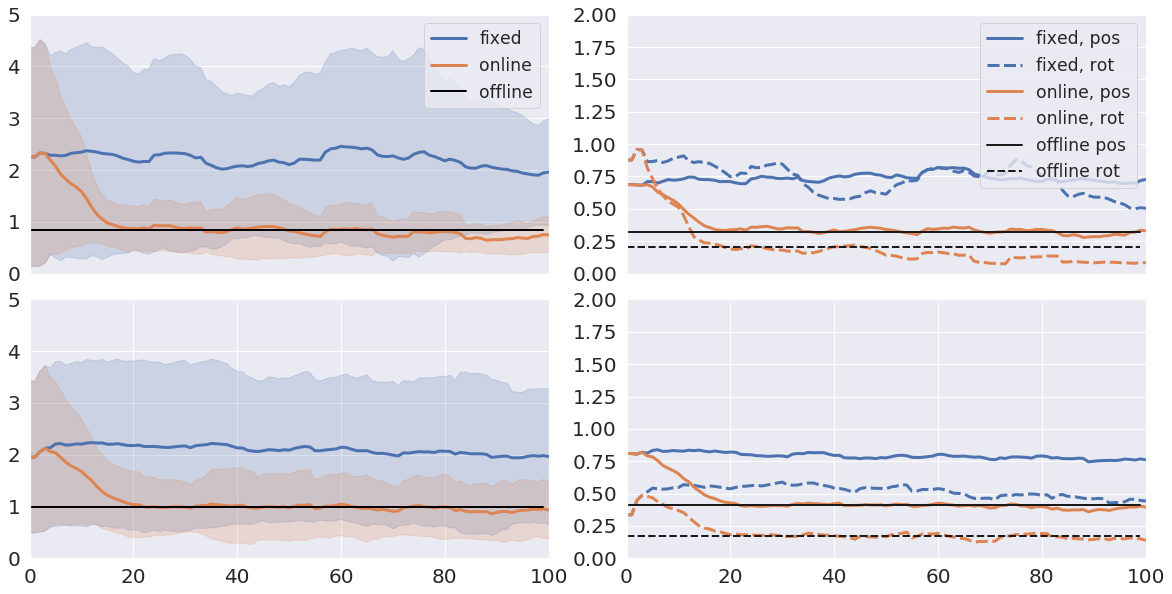}
    \caption{ Exp M2. abs center/abs UR, delrin center/delrin UR}
    \end{figure}
    
    \subsubsection{Exp M3: different objects}
     We consider different objects for online and offline training. All are on abs material.
     
    \begin{figure}
    \centering
    \includegraphics[width=0.48\textwidth]{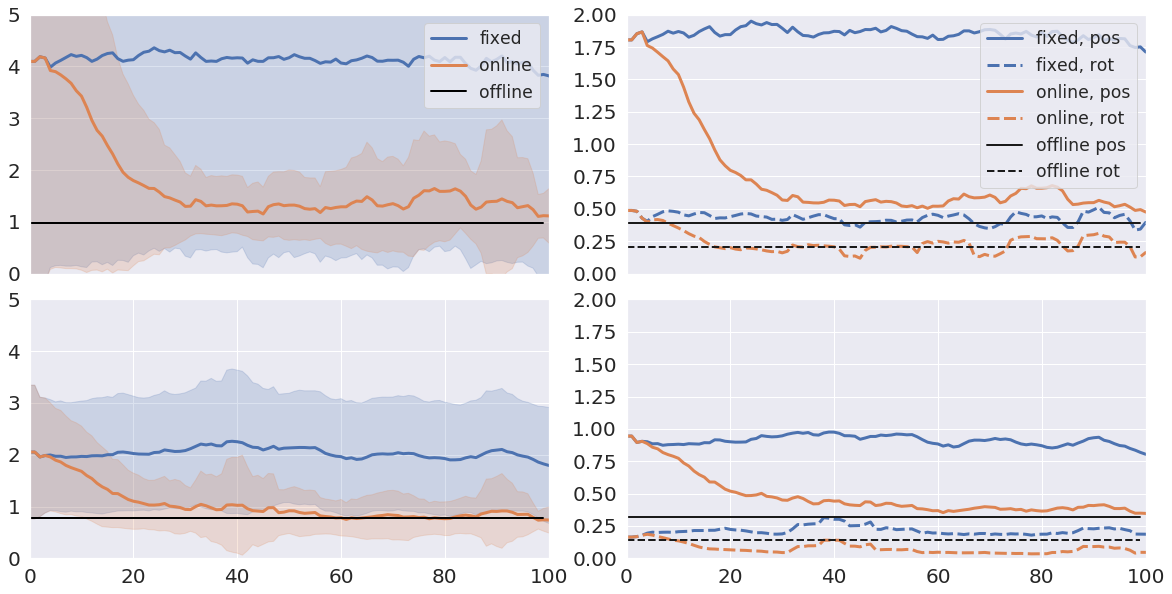}
    \caption{Exp M3. rect1/rect3, rect2/rect3}
    \end{figure}

    \subsubsection{Exp M4: different pushing sides}
    We consider different pushing sides for online and offline training. Object is rect3 and materials are plywood and abs, respectively.
    \begin{figure}
    \centering
    \centering
    \includegraphics[width=0.48\textwidth]{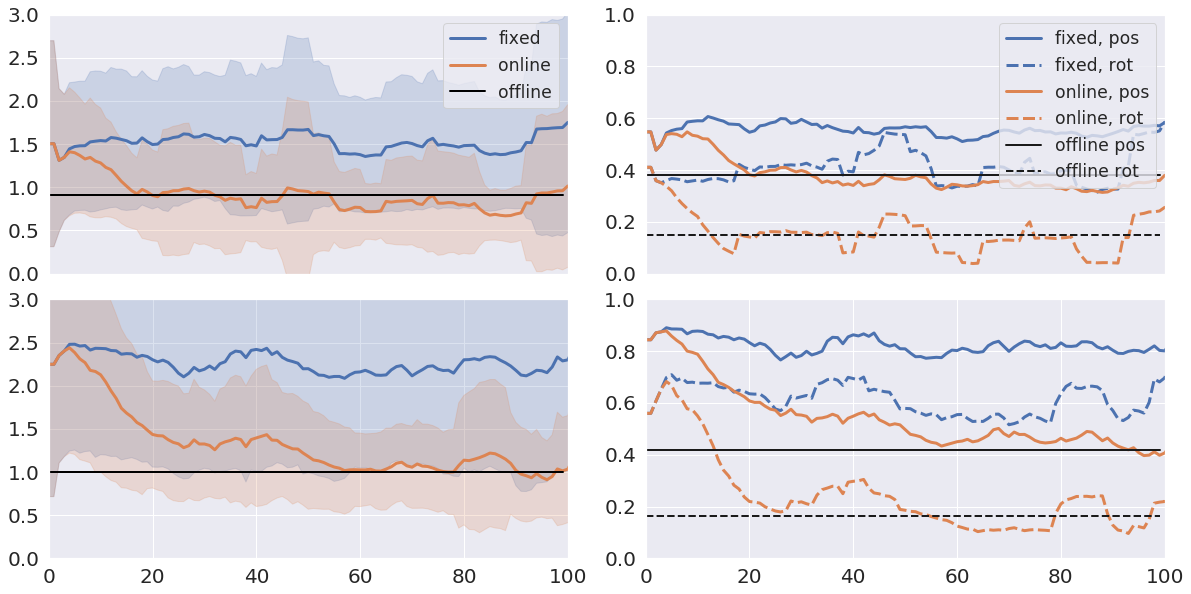}
    \caption{Exp M4. plywood front/plywood left, abs front/abs left}
    \end{figure}
        
    \subsubsection{Exp M5: different objects, materials, pushing sides, COM offsets}
    We consider different objects, materials, pushing sides, COM offsets for offline and online.
    \begin{figure}
    \centering
    \centering
    \includegraphics[width=0.48\textwidth]{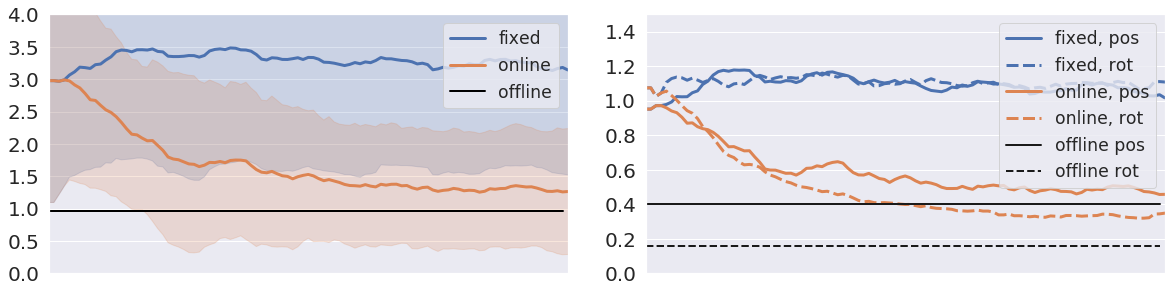}
    \caption{Exp M5. rect2, delrin, front, center/rect3, abs, left, UR
    }
    \end{figure}
   
   %%%%%%%%%%%%%%%%%%%%%%%%%%%%%%%%%%%%%
    \begin{table}
    \centering
        \scriptsize
        \setlength{\tabcolsep}{0.3em} % for the horizontal padding
        \renewcommand{\arraystretch}{1.2}% for the vertical padding
        \begin{tabular}{|p{8mm}||p{7mm}|p{7mm}|p{7mm}|p{7mm}||p{7mm}|p{7mm}|p{7mm}|p{7mm}|}
            \hline
            \multirow{2}{*}{Exps} & \multicolumn{4}{c||}{Positional Losses} & %
                \multicolumn{4}{c|}{Rotational Losses}\\
            \cline{2-9}
             & Offline NN & Offline & Fixed  & Online & Offline NN  & Offline & Fixed & Online
            \\
            \hline
            \multirow{3}{*}{Exp M1} 
             & 0.392 & 0.398 & 0.418 & 0.315 & 0.148 & 0.162 & 0.295 & 0.106
             \\
             \cline{2-9}
             & 0.390 & 0.406 & 0.419 & 0.347 & 0.150 & 0.153 & 0.313 & 0.160
             \\
             \cline{2-9}
             & 0.382& 0.376 & 0.642 & 0.554 & 0.269 & 0.247 & 0.981 & 0.492
             \\
             \hline
            \multirow{2}{*}{Exp M2} 
             & 0.303 & 0.321 & 0.677 & 0.320 & 0.155 & 0.202 & 0.564 & 0.145
             \\
             \cline{2-9}
             & 0.395 & 0.410 & 0.707 & 0.398 & 0.151 & 0.172 & 0.453 & 0.212
             \\
             \hline
            \multirow{2}{*}{Exp M3} 
             & 0.388 & 0.388 & 1.718 & 0.562 & 0.255 & 0.209 & 0.394 & 0.176
             \\
             \cline{2-9}
             & 0.307 & 0.322 & 0.827 & 0.380 & 0.161 & 0.144 & 0.196 & 0.088
             \\
             \hline
             
            \multirow{2}{*}{Exp M4} 
             & 0.383 & 0.381 & 0.607 & 0.364 & 0.148 & 0.148 & 0.480 & 0.176
             \\
             \cline{2-9}
             & 0.425 & 0.419 & 0.828 & 0.477 & 0.167 & 0.164 & 0.670 & 0.212
             \\
             \hline
             
            \multirow{2}{*}{Exp M5} 
             & 0.391 & 0.400 & 0.989 & 0.506 & 0.149 & 0.156 & 0.987 & 0.353
             \\
             \cline{2-9}
             & 0.378 & 0.381 & 0.631 & 0.476 & 0.265 & 0.246 & 0.778 & 0.462
             \\
             \hline

        \end{tabular}
        \caption{Experiments of the MIT push dataset.}
        \label{table:MIT}
    \end{table}

%%%%%%%%%%%%%%%%%%%%%%%%%%%%%%%%%%%%%
\subsection{TurtleBot3 Experiments}
    % \begin{itemize}
    %     \item Two objects: box1 and box2.
    %     \item Two box bottom surface materials: paper (original) and plastic. 
    %     \item Three COM positions: center, upper right corner(UR), lower right corner(LR)
    %     \item Pushing sides: front/right/back/left
    % \end{itemize}

\begin{figure}
    \centering
    \centering
    \includegraphics[width=0.45\textwidth]{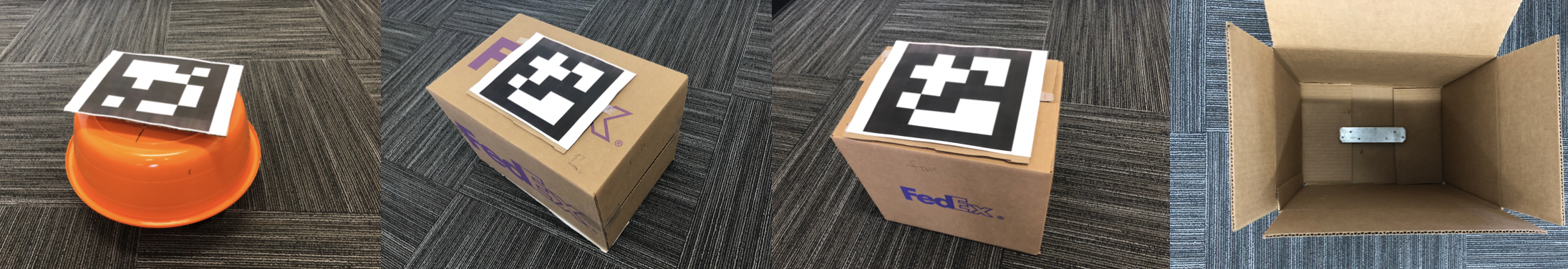}
    \caption{From left to right: TurtleBot3, box1, box2, and an opened box with a stationary object.
    We track the robot and object positions and orientations by their ArUco markers using a ceiling camera. 
    We consider two different surface materials, paper (original) and plastic for the bottom of the boxes, and three COM positions, center, upper right corner (UR), lower right corner (LR), done by moving a stationary object in the box.
    % Pushing sides: front/right/back/left
    }
    \label{fig:turtlebot}
    \end{figure}

    % \monacomment{fill in this} See appendix for specifications.
    See Figure \ref{fig:turtlebot} for our TurtleBot3 setting.
    The experiments are conducted as following. For a certain object, material, COM and pushing side setting, a total of nine straight push trajectories are used to test online learning. These trajectories are a combination of different pushing points and pushing angles. For offline training, each setting has 27 straight push trajectories. Each offline and online trajectory contains around 100 data points, with a data collection frequency of 4 Hz. In order to make a reasonable prediction, we set the prediction horizon to be 2 seconds, predicting the object's position and rotation 2 seconds from the current time frame. 
    
    \subsubsection{Exp R1: different materials}
        We consider box1 and do offline and online training with different materials.
    \begin{figure}
    \centering
    \centering
    \includegraphics[width=0.48\textwidth]{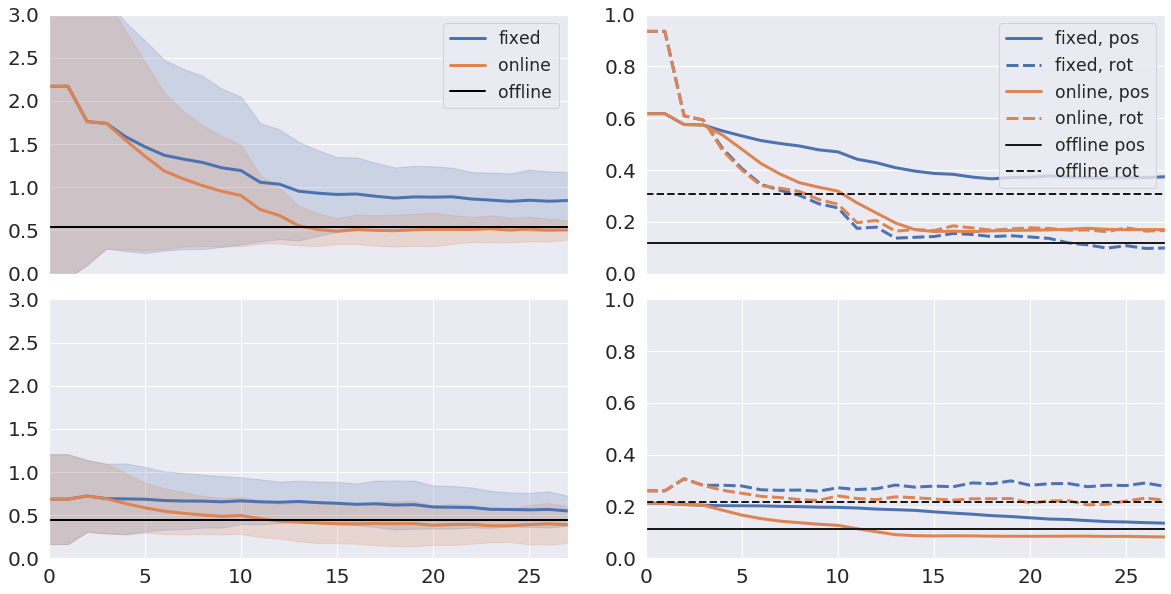}
    \caption{Exp R1. paper/plastic, plastic/paper
    }
    \end{figure}
   
    \subsubsection{Exp R2: different COMs}
        We train with different COM settings online with either paper or plastic. Note that in two EXP R1 and R2 settings the online loss for the fixed model also decreases. This actually is not due to learning, but because of the rotational loss resulting from the large initial rotational movement in the two cases.

    \begin{figure}
    \centering
    \centering
    \includegraphics[width=0.48\textwidth]{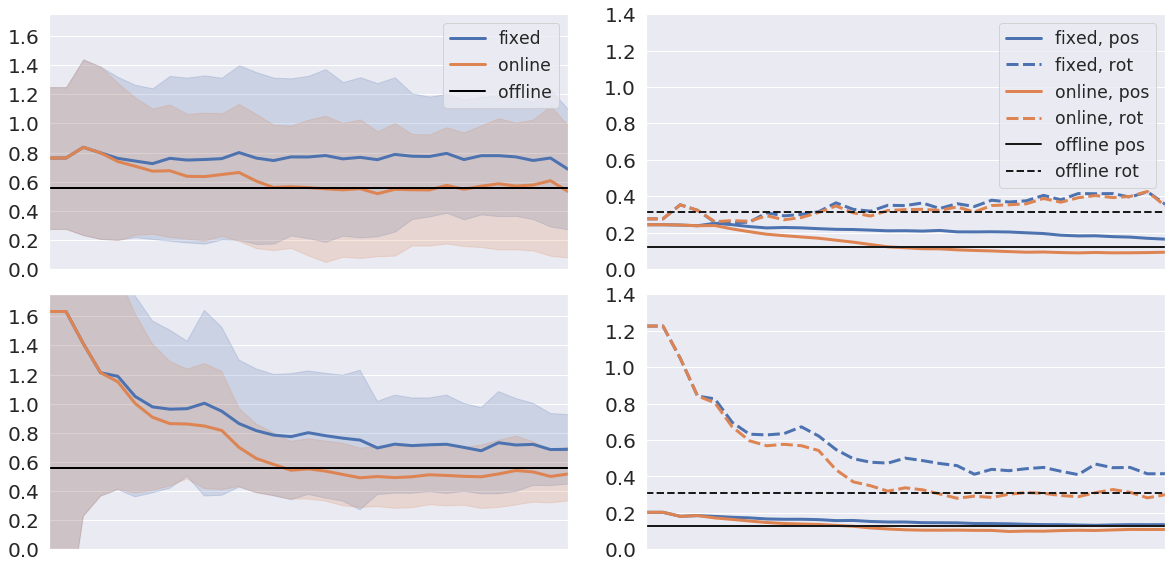}
    \caption{Exp R2. paper center/paper LR, paper center/paper UR
    }
    \end{figure}
    
    \subsubsection{Exp R3: different objects}
        We consider different objects for online and offline training on either paper or plastic.
    \begin{figure}
    \centering
    \centering
    \includegraphics[width=0.48\textwidth]{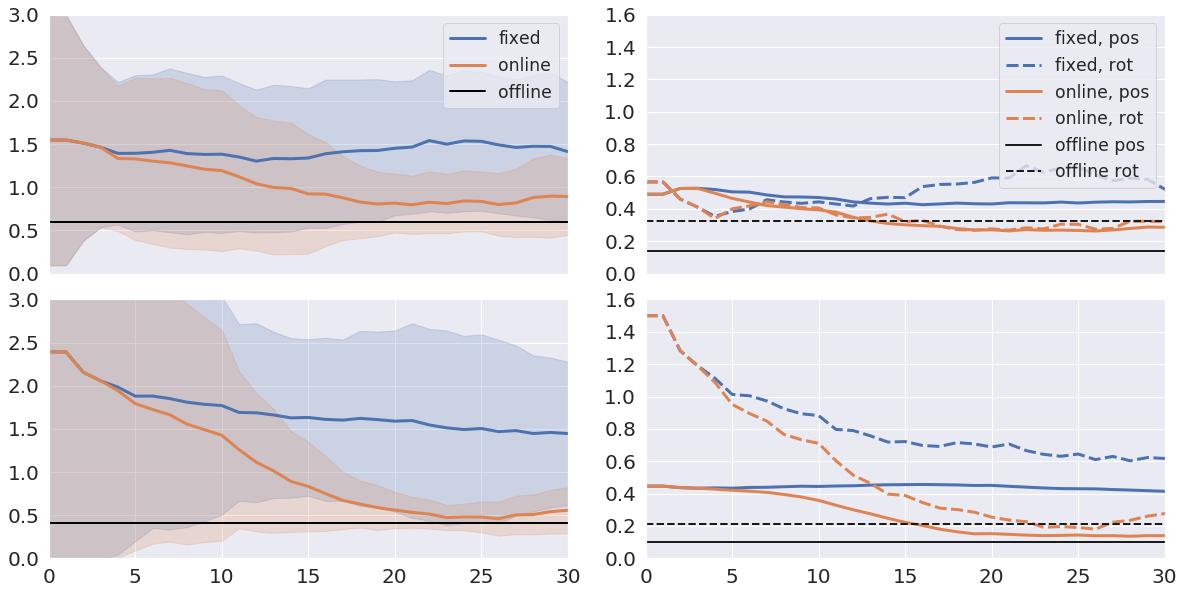}
    \caption{Exp R3. paper box1/box2, Exp R3. plastic box1/box2
    }
    \end{figure}
        
    \subsubsection{Exp R4: different pushing sides}
         We consider different pushing sides for online and offline training with box1.
    \begin{figure}[h]
    \centering
    \centering
    \includegraphics[width=0.48\textwidth]{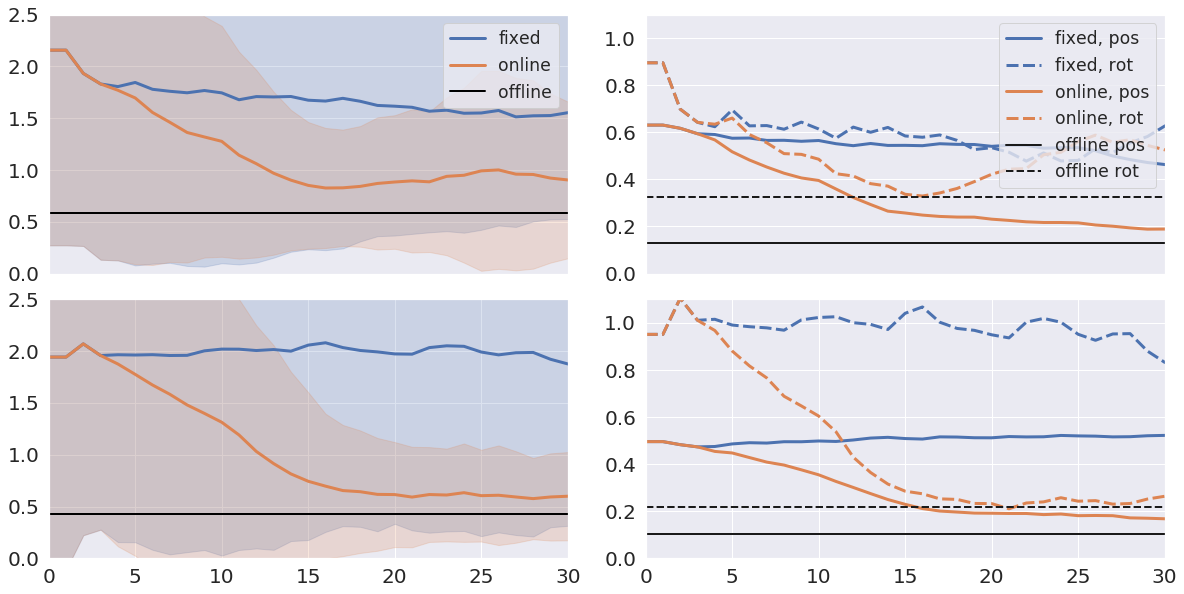}
    \caption{Exp R4. paper front/paper right, plastic front/plastic right}
    \end{figure}
    
    \subsubsection{Exp R5: all different}
    See the online prediction video.

    \begin{table}
        \centering
        \scriptsize
        \setlength{\tabcolsep}{0.3em} % for the horizontal padding
        \renewcommand{\arraystretch}{1.2}% for the vertical padding
        \begin{tabular}{|p{8mm}||p{7mm}|p{7mm}|p{7mm}|p{7mm}||p{7mm}|p{7mm}|p{7mm}|p{7mm}|}
            \hline
            \multirow{2}{*}{Exps} & \multicolumn{4}{c||}{Positional Losses} & %
                \multicolumn{4}{c|}{Rotational Losses}\\
            \cline{2-9}
             & Offline NN & Offline & Fixed & Online & Offline NN & Offline &  Fixed & Online 
            \\
            \hline
            \multirow{3}{*}{Exp R1} 
              & 0.116 & 0.117 & 0.441 & 0.287 & 0.319 & 0.307 & 0.264 & 0.292
             \\
             \cline{2-9}
             & 0.105 & 0.112 & 0.176 & 0.118 & 0.183 & 0.219 & 0.279 & 0.236
             \\
             \cline{2-9}
             & 0.382& 0.376 & 0.642 & 0.554 & 0.269 & 0.247 & 0.981 & 0.492
             \\
             \hline
            \multirow{3}{*}{Exp R2} 
             & 0.160 & 0.120 & 0.186 & 0.123 & 0.320 & 0.316 & 0.331 & 0.321
             \\
             \cline{2-9}
             & 0.138 & 0.126 & 0.145 & 0.121 & 0.282 & 0.307 & 0.523 & 0.435
             \\
             \cline{2-9}
             & 0.102 & 0.109 & 0.190 & 0.148 & 0.207 & 0.194 & 0.268 & 0.201
             \\
             \hline
            \multirow{2}{*}{Exp R3} 
             & 0.118 & 0.138 & 0.449 & 0.316 & 0.281 & 0.325 & 0.453 & 0.312
             \\
             \cline{2-9}
             & 0.098 & 0.099 & 0.419 & 0.213 & 0.178 & 0.212 & 0.701 & 0.455
             \\
             \hline
            \multirow{2}{*}{Exp R4} 
             & 0.123 & 0.128 & 0.483 & 0.269 & 0.291 & 0.327 & 0.603 & 0.498
             \\
             \cline{2-9}
             & 0.105 & 0.104 & 0.490 & 0.232 & 0.182 & 0.218 & 0.924 & 0.388
             \\
             \hline

        \end{tabular}
        \caption{Experiments by TurtleBot3.}
        \label{table:turtlebot}
    \end{table}

%%%%%%%%%%%%%%%%%%%%%%%%%%%%%%%%%%%%%%%%%%%%%%%%%%%%%%%%%%%%%%%%%%%%%%%%%%%%%%%%

\section{Conclusion}

In this paper, we propose a combined prediction model and an online framework for planar push prediction.
Our method takes advantages of both offline and online learning by utilizing a neural network module and several analytical components.
We conducted two sets of real-robot pushing experiments to verify the prediction accuracy and online adaptation ability of our model.
The experiments showed that our model can achieve similar offline performance as a pure neural network model.
In situations different from offline data, with online learning our combined prediction model is able to reduce the loss to a level similar to the offline loss, as well as reducing the uncertainty in both positional and rotational predictions.

%%%%%%%%%%%%%%%%%%%%%%%%%%%%%%%%%%%%%%%%%%%%%%%%%%%%%%%%%%%%%%%%%%%%%%%%%%%%%%%%

% \section*{ACKNOWLEDGMENT}

% \cite{IEEEexample:IEEEwebsite}
\newpage
\bibliographystyle{IEEEtran}
\bibliography{IEEEabrv,references}

\end{document}